\documentclass[conference]{IEEEtran}
\IEEEoverridecommandlockouts
\usepackage{cite}
\usepackage{amsmath,amssymb,amsfonts}
\usepackage{algorithmic}
\usepackage{graphicx}
\usepackage{textcomp}
\usepackage{xcolor}
\usepackage{csquotes}
\usepackage{siunitx}
\usepackage{subcaption}
\usepackage{gensymb}
\usepackage{url}
\usepackage{multirow}
\usepackage{colortbl}
\usepackage{hyperref}
\usepackage{cleveref}

\def\BibTeX{{\rm B\kern-.05em{\sc i\kern-.025em b}\kern-.08em
    T\kern-.1667em\lower.7ex\hbox{E}\kern-.125emX}}
\begin{document}

\title{\LARGE \bf
Automatic Calibration for an Open-source Magnetic Tactile Sensor
}

\author{Lowiek Van den Stockt$^{1*}$, Remko Proesmans$^{1*}$ and Francis wyffels$^{1}$
\thanks{$^{1}$Lowiek Van den Stockt, Remko Proesmans and Francis wyffels are with the IDLab-AIRO research lab at Ghent University -- imec, Technologiepark-Zwijnaarde
126, 9052 Zwijnaarde, Belgium. Remko Proesmans is a predoctoral fellow of the Research Foundation Flanders (FWO) under grant agreement no. 1S15923N. This work was also partially supported by the euROBIn Project (EU grant number 101070596). {\tt\footnotesize remko.proesmans@ugent.be}}
\thanks{$^*$Lowiek Van den Stockt and Remko Proesmans made equal contributions to this work.}%
}

\maketitle

\begin{abstract}
Tactile sensing can enable robots to perform complex, contact-rich tasks. 
Magnetic sensors offer accurate three-axis force measurements while using affordable materials. 
Calibrating such a sensor involves either manual data collection, or automated procedures with precise mounting of the sensor relative to an actuator. 
We present an open-source magnetic tactile sensor with an automatic, in situ, gripper-agnostic calibration method, after which the sensor is immediately ready for use. 
Our goal is to lower the barrier to entry for tactile sensing, fostering collaboration in robotics.
Design files and readout code can be found at
\href{https://github.com/LowiekVDS/Open-source-Magnetic-Tactile-Sensor}{https://github.com/LowiekVDS/Open-source-Magnetic-Tactile-Sensor}.

\end{abstract}


\section{Introduction}
Tactile sensing in robotics is crucial for contact-rich tasks like food picking \cite{yagawa2023} and in-hand manipulation \cite{yin2023rotating}. 
Widespread sensor modalities include piezoresistive, capacitive, barometric, optical and magnetic sensing.
Among these, magnetic sensors can be made with readily available materials and components while providing accurate three-axis force measurements  \cite{uskin}.
They work by suspending small magnets \cite{uskin, kristanto2019, sathe2023, dai2022, lesignor2022}, coils \cite{xie2021}, or distributed magnetic substrates \cite{yan2021, reskin} above a grid of Hall sensors. 

Typically, a data-driven calibration procedure is performed to interpret the data from such sensors.
In \cite{uskin, kristanto2019, sathe2023}, a human presses the magnetic sensor against a force/torque (F/T) sensor to capture calibration data.
Other work proposes automated procedures: \cite{reskin, xie2021, yan2021} use a robot equipped with an F/T sensor to press down on the magnetic sensor in a predefined grid, whereas \cite{lesignor2022} has the magnetic sensor pressing down on a load cell mounted to a moving platform.

Schmitz et al. \cite{uskin, kristanto2019, sathe2023} have commercialised three-axis tactile fingertips based on magnetic sensing, greatly incentivising further research in tactile sensing. 
However, the monetary cost can still be a hurdle for other groups to adopt these sensors. 
In the spirit of democratising tactile sensors, we are developing open-source three-axis force sensors: here, we present a baseline fingertip  with a 2-by-2 taxel grid.
To make adoption in other research groups as straightforward as possible, we propose an automatic calibration procedure with the sensor already mounted to the robot that will make use of it.
This way, the sensor is ready for use immediately after calibration, as opposed to previous work \cite{lesignor2022, reskin, xie2021, yan2021}.

\section{Sensor design} \label{ss:sensor_design}

\begin{figure}[tpb]
  \centering
  \includegraphics[width=0.8\linewidth]{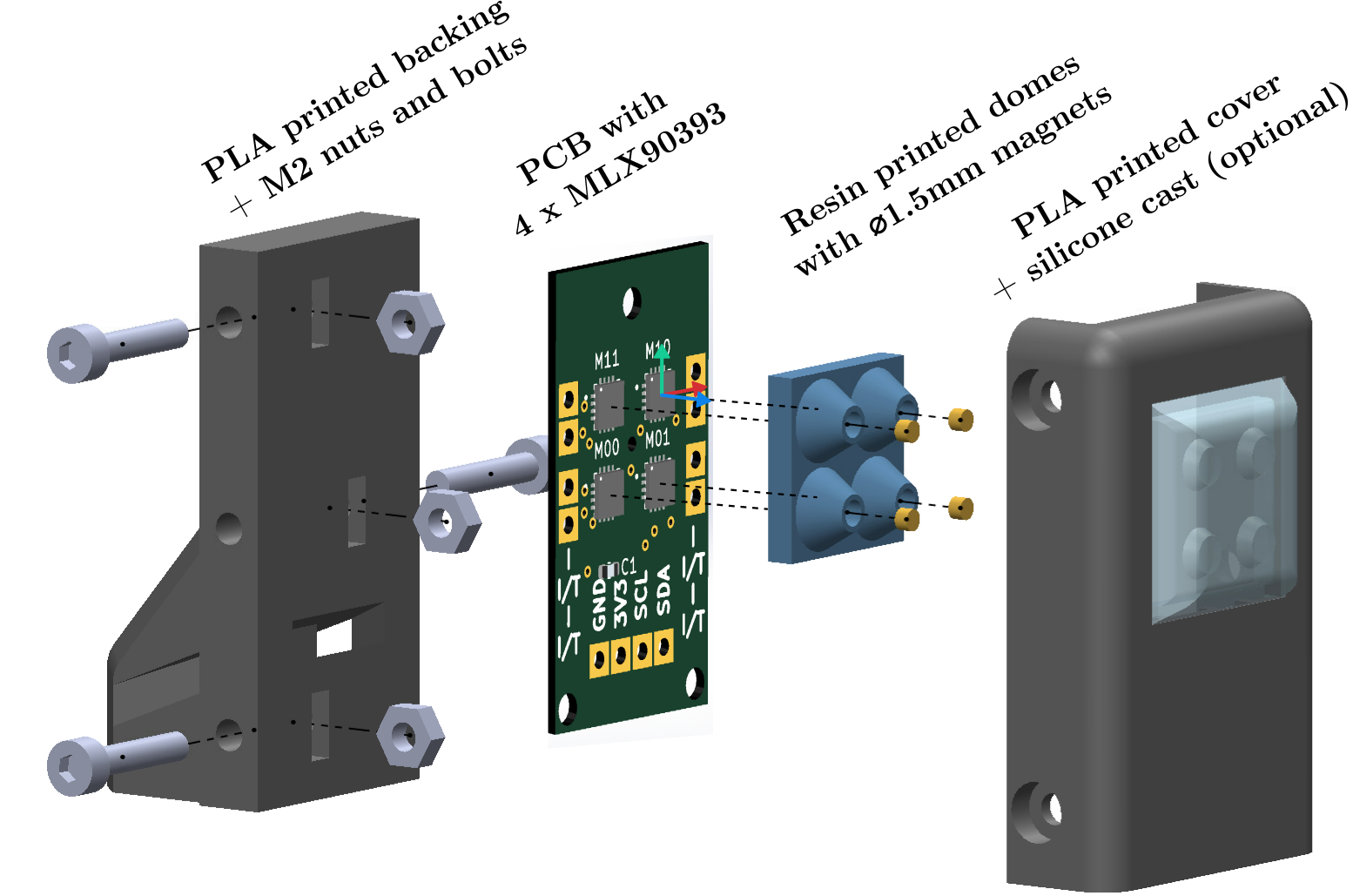}
  \caption{Fingertip structure. The magnets are glued into the domes, the domes are glued onto the PCB. The PCB is pressed between the cover and backing, and the cover is bolted to the backing.}
  \label{fig:exploded}
\end{figure}

\begin{figure}[tpb]
  \centering
  \includegraphics[width=0.8\linewidth]{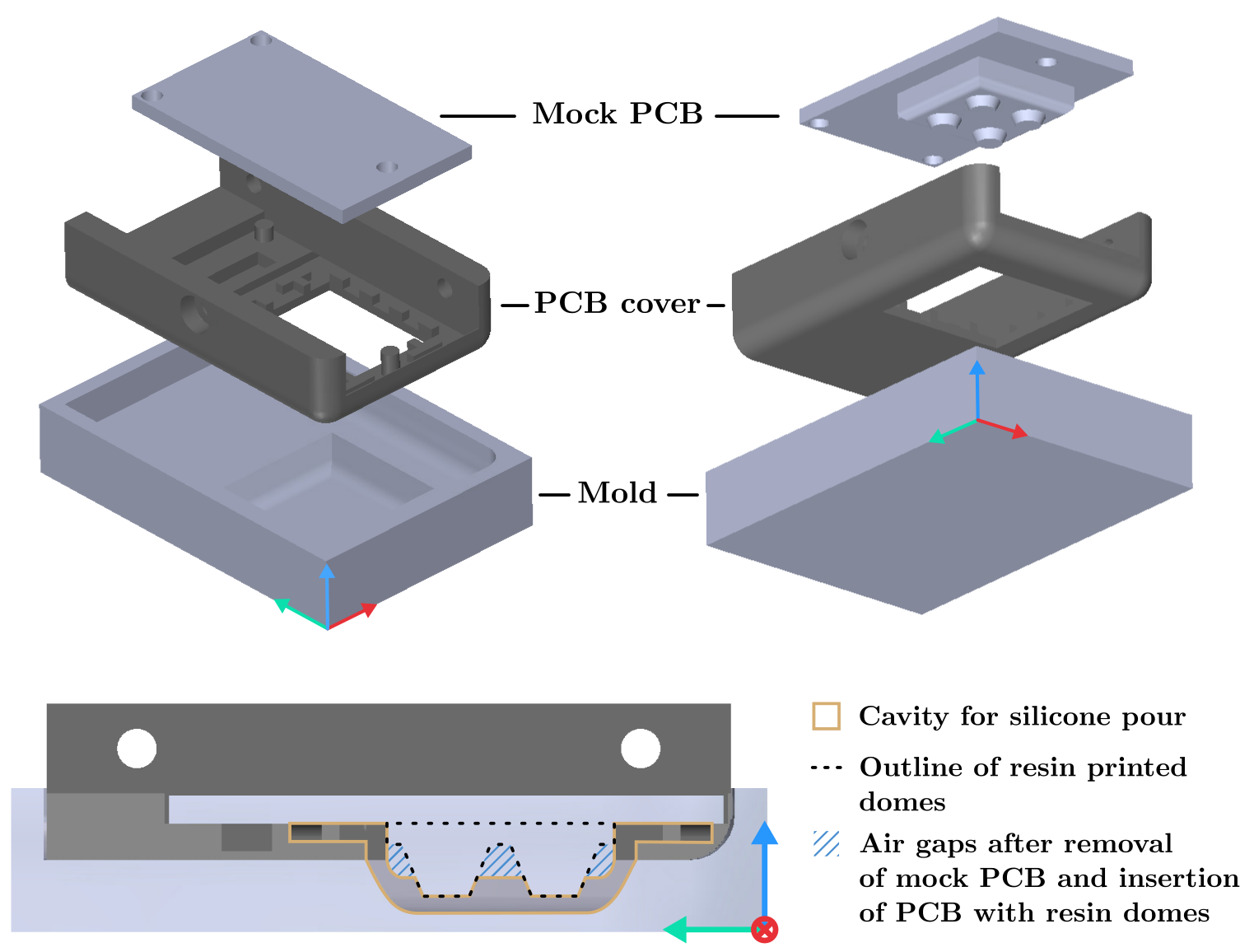}
  \caption{Mould for silicone pour. The domes on the mock PCB are shorter than the resin-printed domes, such that after removal of the mock PCB and insertion of the PCB with resin domes, an air cavity is present all around each dome.}
  \label{fig:mold}
\end{figure}

At the heart of the fingertip design, shown in Fig.~\ref{fig:exploded}, lies a printed circuit board (PCB) with a 2-by-2 grid of MLX90393 Hall effect sensors connected to an I$^2$C bus.
A dome structure similar to \cite{uskin, sathe2023} is printed in Flexible FLGR02 resin 
using a Formlabs Form~2 printer.
Cylindrical magnets with a height of \SI{1}{\milli\metre} and diameter of \SI{1.5}{\milli\metre} are superglued into the domes, and the dome structure itself is glued to the PCB.
A cover for the PCB is 3D printed in PLA using a Prusa\,i3~MK3.
Optionally, a smooth contact surface for grasping can be cast onto the cover using Silicone Addition Colorless 50 by Silicones and More.
Fig.~\ref{fig:mold} shows the mould used.
Crucially, a mock PCB is inserted into the cover during curing.
The domes on the mock PCB are shorter than the resin domes.
This way, when the mock PCB is removed and the resin domes are inserted after curing, an air gap surrounds each dome.
The air gap ensures that the force applied to the silicone surface is transferred to the resin domes.
Finally, the cover is bolted to a PLA backing (Fig.~\ref{fig:exploded}).
The backing is designed to fit a Robotiq 2F\nobreakdash-85 or 2F\nobreakdash-140 gripper, but we also provide a coupling to fit Robotiq fingertips to Schunk EGU and EGK grippers. 


\section{Automatic Calibration}

\subsection{Hardware Setup}
\begin{figure}[tpb]
  \centering
  \includegraphics[width=0.8\linewidth]{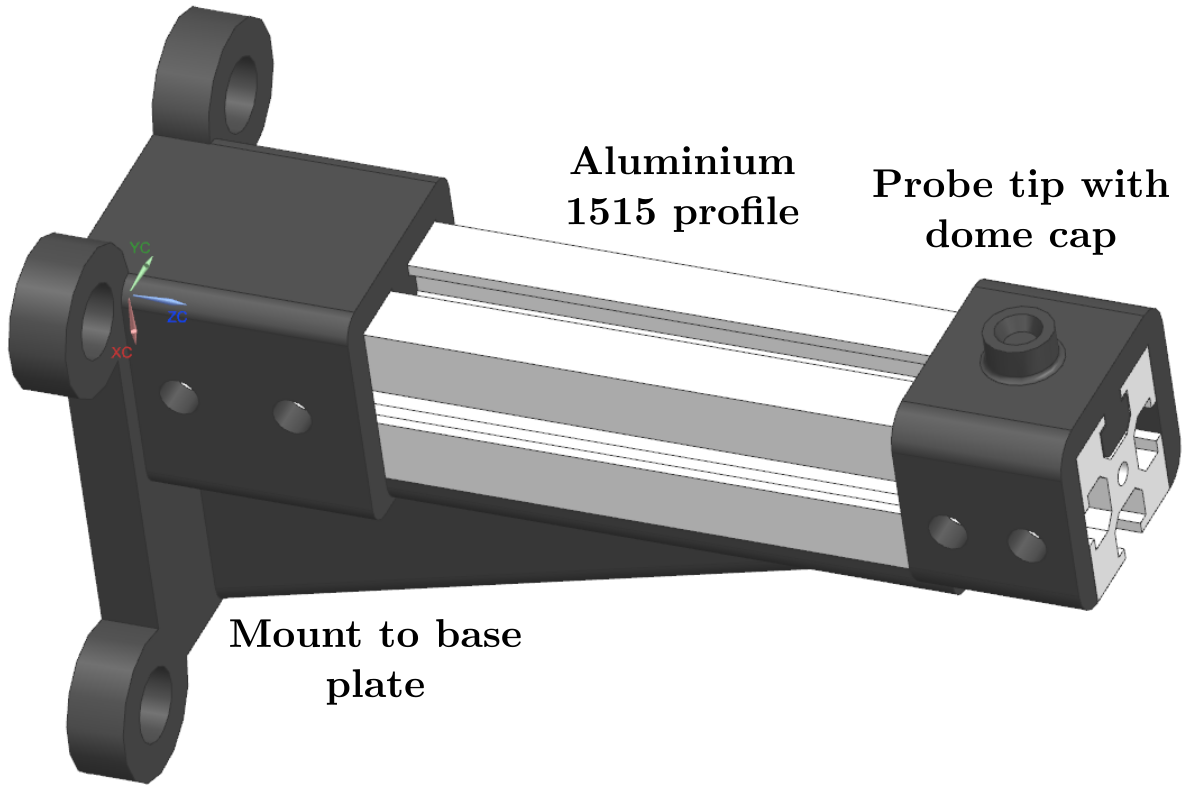}
  \caption{Calibration probe assembly. The probe tip perfectly fits a taxel dome.}
  \label{fig:probe}
\end{figure}

The robotic arm on which the gripper is mounted should feature an F/T sensor.
In this work, we used a Schunk~EGK~40\nobreakdash-MB\nobreakdash-M\nobreakdash-B and a UR3e collaborative arm.
A custom probe, see Fig.~\ref{fig:probe}, is mounted externally to the robot.
The probe is constructed from an aluminium 1515 profile with D-shaped slots and several 3D-printed PLA parts. 
The probe tip is shaped such that it fits on top of one of the resin domes. 
It is important that both the probe tip and the mounting of the probe to the table are rigid.

\subsection{Calibration Methodology}
Calibration is done in situ, meaning the fingertip is mounted on the robot that will make use of it.
At first, the cover (see Fig.~\ref{fig:exploded}) should not have a silicone pad.
If desired, the cover can easily be replaced after calibration without removing the fingertip from the gripper.
The procedure is gripper-agnostic. 
Only the direction of the probe tip expressed in the base frame of the robot should be known.

First, the robot is placed in the vicinity of the sensor, with the fingertip facing the probe tip. 
The robot then touches the probe with the front, the side and the top of the fingertip. 
Contact is observed from the F/T readings.
Knowing the dimensions of the fingertip, the probe can be localised.
The robot subsequently presses one of the resin domes into the probe and exerts a predefined set of forces, see the green annotations in Fig.~\ref{fig:results}, while recording both the F/T and the Hall sensor readings.
This is repeated for each taxel.
The F/T data is averaged over a moving window of 100 samples or \SI{240}{\milli\second}. 
A linear interpolation is applied to account for different readout frequencies of the Hall and F/T sensors. 

Polynomial features up to the third degree are extracted from the Hall readings. 
An 80\%/20\% train-test split is randomly sampled and a least squares linear regressor is fitted on the train set. 
The result is a model per taxel that takes in the X, Y and Z components of the magnetic field and predicts the three components of the force ($F_x$, $F_y$, $F_z$) applied to the taxel.

\section{Results}
\begin{figure}[tpb]
  \centering
  \includegraphics[width=\linewidth]{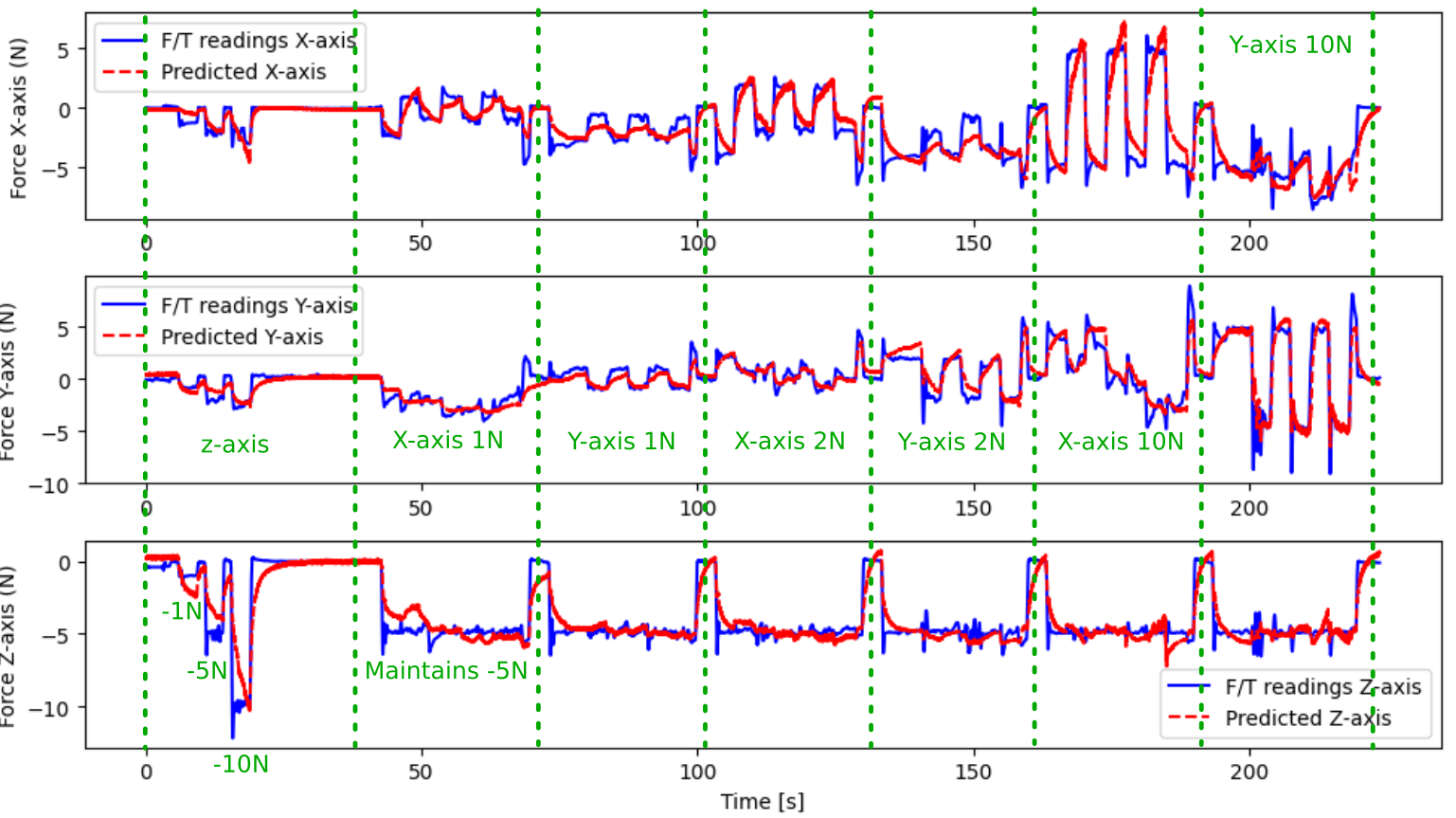}
  \caption{F/T Readings and model predictions for taxel 3.}
  \label{fig:results}
\end{figure}

For each taxel, Table~\ref{results_table} gives the coefficient of determination ($R^2$) and the mean squared error (MSE) of the model when applied to the test set.
Fig.~\ref{fig:results} additionally shows both the true and  predicted forces for taxel 3.
We note that the predicted force curves rise comparatively slowly to the ground truth. 
We believe this is due to the mechanical hysteresis of the resin-printed domes.

\begin{table}[h]
    \centering
    \caption{Test scores for each taxel}
    \begin{tabular}{ccc}
         \hline
         Taxel & $R^2$ & MSE [N$^2$]  \\
         \hline
         0 & 0.81162 & 1.10555 \\
         1 & 0.80103 & 0.97162 \\
         2 & 0.84808 & 0.88912 \\
         3 & 0.84431 & 0.92823 \\
         \hline
         Average & 0.82626 & 0.97363
    \end{tabular}
    \label{results_table}
\end{table}

\section{Conclusion}

We presented an open-source design for a magnetic tactile fingertip and have developed an automatic, gripper-agnostic, in situ calibration strategy.
The calibration method only requires a probe to be mounted along a known direction w.r.t. the base frame of the robot. A polynomial model trained on the magnetic sensor data shows an average $R^2$ of 0.83 and an average MSE of 0.97\,N$^2$. 
In future work, we will open-source a modular fingertip design featuring tactile surfaces with 4\nobreakdash-by\nobreakdash-8, 1\nobreakdash-by\nobreakdash-4, and 2\nobreakdash-by\nobreakdash-2 taxel grids.
In addition, we aim to compensate for hysteresis and to examine a second calibration stage that can account for mechanical and magnetic coupling. 
Our goal is to lower the barrier to entry of tactile sensors and encourage their adoption across robotics labs.

\bibliographystyle{IEEEtran}
\bibliography{references.bib}

\begin{thebibliography}{10}
\providecommand{\url}[1]{#1}
\csname url@samestyle\endcsname
\providecommand{\newblock}{\relax}
\providecommand{\bibinfo}[2]{#2}
\providecommand{\BIBentrySTDinterwordspacing}{\spaceskip=0pt\relax}
\providecommand{\BIBentryALTinterwordstretchfactor}{4}
\providecommand{\BIBentryALTinterwordspacing}{\spaceskip=\fontdimen2\font plus
\BIBentryALTinterwordstretchfactor\fontdimen3\font minus \fontdimen4\font\relax}
\providecommand{\BIBforeignlanguage}[2]{{%
\expandafter\ifx\csname l@#1\endcsname\relax
\typeout{** WARNING: IEEEtran.bst: No hyphenation pattern has been}%
\typeout{** loaded for the language `#1'. Using the pattern for}%
\typeout{** the default language instead.}%
\else
\language=\csname l@#1\endcsname
\fi
#2}}
\providecommand{\BIBdecl}{\relax}
\BIBdecl

\bibitem{yagawa2023}
R.~Yagawa, R.~Ishikawa, M.~Hamaya, K.~Tanaka, A.~Hashimoto, and H.~Saito, ``Learning food picking without food: Fracture anticipation by breaking reusable fragile objects,'' in \emph{2023 IEEE International Conference on Robotics and Automation (ICRA)}, 2023, pp. 917--923.

\bibitem{yin2023rotating}
Z.-H. Yin, B.~Huang, Y.~Qin, Q.~Chen, and X.~Wang, ``Rotating without seeing: Towards in-hand dexterity through touch,'' 2023.

\bibitem{uskin}
T.~P. Tomo, M.~Regoli, A.~Schmitz, L.~Natale, H.~Kristanto, S.~Somlor, L.~Jamone, G.~Metta, and S.~Sugano, ``A new silicone structure for uskin—a soft, distributed, digital 3-axis skin sensor and its integration on the humanoid robot icub,'' \emph{IEEE Robotics and Automation Letters}, vol.~3, no.~3, pp. 2584--2591, 2018.

\bibitem{kristanto2019}
H.~Kristanto, P.~Sathe, A.~Schmitz, C.~Hsu, T.~P. Tomo, S.~Somlor, and S.~Sugano, ``Development of a 3-axis human fingertip tactile sensor based on distributed hall effect sensors,'' in \emph{2019 IEEE-RAS 19th International Conference on Humanoid Robots (Humanoids)}, 2019, pp. 1--7.

\bibitem{sathe2023}
P.~Sathe, A.~Schmitz, T.~P. Tomo, S.~Somlor, S.~Funabashi, and S.~Shigeki, ``Fingertac - an interchangeable and wearable tactile sensor for the fingertips of human and robot hands,'' in \emph{2023 IEEE/RSJ International Conference on Intelligent Robots and Systems (IROS)}, 2023, pp. 10\,813--10\,820.

\bibitem{dai2022}
K.~Dai, X.~Wang, A.~Rojas, E.~Harber, Y.~Tian, N.~Paiva, J.~Gnehm, E.~Schindewolf, H.~Choset, V.~Webster-Wood, and L.~Li, ``Design of a biomimetic tactile sensor for material classification,'' in \emph{2022 International Conference on Robotics and Automation (ICRA)}, 2022, pp. 10\,774--10\,780.

\bibitem{lesignor2022}
T.~Le~Signor, N.~Dupré, and G.~F. Close, ``A gradiometric magnetic force sensor immune to stray magnetic fields for robotic hands and grippers,'' \emph{IEEE Robotics and Automation Letters}, vol.~7, no.~2, pp. 3070--3076, 2022.

\bibitem{xie2021}
S.~Xie, Y.~Zhang, M.~Jin, C.~Li, and Q.~Meng, ``High sensitivity and wide range soft magnetic tactile sensor based on electromagnetic induction,'' \emph{IEEE Sensors Journal}, vol.~21, no.~3, pp. 2757--2766, 2021.

\bibitem{yan2021}
Y.~Yan, Z.~Hu, Z.~Yang, W.~Yuan, C.~Song, J.~Pan, and Y.~Shen, ``Soft magnetic skin for super-resolution tactile sensing with force self-decoupling,'' \emph{Science Robotics}, vol.~6, no.~51, p. eabc8801, 2021.

\bibitem{reskin}
R.~Bhirangi, T.~Hellebrekers, C.~Majidi, and A.~Gupta, ``Reskin: versatile, replaceable, lasting tactile skins,'' in \emph{2021 Conference on Robotic Learning (CoRL)}, 2021.

\end{thebibliography}

\end{document}